# FD-GAN: Face de-morphing generative adversarial network for restoring accomplice's facial image

Fei Peng[1], Member, IEEE, Le-Bing Zhang[1], and Min Long[2]
[1]College of Computer Science and Electronic Engineering, Hunan University, 410082 Changsha, China
[2]College of Computer and Communication Engineering, Changsha University of Science and Technology, 410114 Changsha, China

Corresponding author: Fei Peng (e-mail: eepengf@gmail.com).

**ABSTRACT** Face morphing attack is proved to be a serious threat to the existing face recognition systems. Although a few face morphing detection methods have been put forward, the face morphing accomplice's facial restoration remains a challenging problem. In this paper, a face de-morphing generative adversarial network (FD-GAN) is proposed to restore the accomplice's facial image. It utilizes a symmetric dual network architecture and two levels of restoration losses to separate the identity feature of the morphing accomplice. By exploiting the captured facial image (containing the criminal's identity) from the face recognition system and the morphed image stored in the e-passport system (containing both criminal and accomplice's identities), the FD-GAN can effectively restore the accomplice's facial image. Experimental results and analysis demonstrate the effectiveness of the proposed scheme. It has great potential to be implemented for detecting the face morphing accomplice in a real identity verification scenario.

## I. INTRODUCTION

With the development of face biometrics, face recognition systems (FRS) with high accuracy are extensively used in our daily life. In real applications, a prevalently used FRS is Automatic Border Control (ABC) system. It can expediently verify a person's identity with one's electronic machine readable travel document (eMRTD) [1], which contains a facial reference image. In the past twelve years, more than 800 million eMRTD instances have been issued, following the specifications from the International Civil Aviation Organization (ICAO).

Recently, it was found that a novel identity theft scenario can easily spoof the FRS of ABC [2]. The idea of the attack is: a morphed facial image (a combination of two or more real facial images) is first generated, and it looks like multiple real persons (e.g. a criminal and an accomplice). Then the morphed image is enrolled as an identity template of the FRS. In a successful attack, both the criminal and the accomplice can match the template stored in the FRS. It means that a wanted criminal is able to obtain a legitimate e-passport or eMRTD by morphing his facial image with an accomplice. Some samples of the morphed facial images are shown in Figure 1.

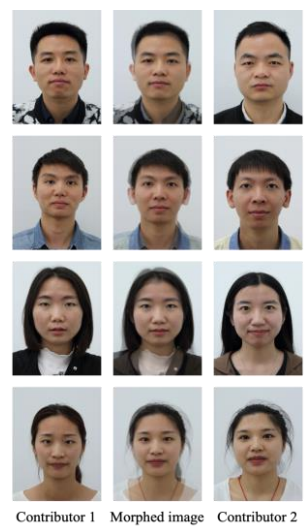

**FIGURE 1.** Example of the morphed images generated from two contributors.

After this idea was put forward, some concerns about the vulnerability of commercial FRS with respect to morphed face attack have been investigated [3-7]. It was proved that face morphing attack is posing a serious threat to the existing FRS. To countermeasure this situation, some face morphing detection methods have been proposed to protect the security of FRS [8-16]. However, all of these methods only detected the presence of face morphing attack, the restoration of the accomplice's facial image has not been seriously discussed.

Recently, a face de-morphing method was proposed in [17]. It restored the facial image of the morphing accomplice by reversing the face morphing operation. Unfortunately, it needs the prior knowledge on the morphing parameters of the face morphing, which is impractical in real applications.

Inspired by the popular use of generative adversarial networks (GAN) [18] in image synthesis, we consider the restoration of the accomplice's facial image from a perspective of a learning-based generation. Different from the face de-morphing by inverting the morphing process in [17], our goal is to obtain the accomplice's identity information in the morphed facial image without the prior knowledge of morphed face generation. To accomplish this, FD-GAN is designed, and it utilizes a symmetric dual network architecture and two levels of restoration losses to extract the accomplice's identity feature, and restore the facial image of the accomplice. The main contributions are in the following.

- FD-GAN is proposed to restore the accomplice's facial image without prior knowledge of the morphed facial images. To the best of our knowledge, it is the first attempt to exploit learning-based generation approach for facial image restoration in face morphing detection.
- A symmetric dual network architecture, pixel-level and feature-level restoration losses of FD-GAN are designed to disentangle the identity features of the morphing participants hidden in the morphed facial image.
- Experimental results and analysis illustrate it can effectively restore the facial image of the morphing accomplice.

The rest of the paper is organized as follows. The related work is introduced in Section II. The proposed FD-GAN is described in Section III. Experimental results are provided in Section IV. Finally, some conclusions are drawn in Section V.

## II. RELATED WORK

Currently, the researches related to face morphing are mainly concentrated on the vulnerability of FRS to face morphing attack and the face morphing detection.

### A. VULNERABILITY OF FRS TO FACE MORPHING ATTACK

Face morphing attack was first introduced in [2] by M. Ferrera *et al*. With a morphed facial image, which is similar to the appearance of multiple persons, it can match with multiple persons in FRS. Meanwhile, it indicates that a wanted criminal may use a morphed image, which is composited by his own facial image and an accomplice's facial image, to acquire an official authority "forged" ID card. However, the morphed facial images are manually generated, which is time-consuming and not suitable for generating large amounts of morphing images to verify the vulnerability of FRS. After that, an automatic generation of visually faultless facial morphing image was proposed in [9]. With this technique, a large number of morphed facial images (including complete morphing and splicing morphing images) can be generated. Furthermore, the quality of the morphed images is verified by human observers (including face recognition experts) and a commercial-off-the-shelf FRS (Luxand FaceSDK 6.1).

Recently, D. J. Robertson *et al*. examined the potential route of using facial morphing in fraudulent documents, and found that it is possible to forge identity by morphing facial images [3]. Meanwhile, M. Gomez-Barrero *et al*. also confirmed that not only face recognition system but also iris and fingerprint recognition systems are vulnerable to morphing attack [4], and a framework to evaluate the vulnerability of biometric systems to morphing attack is proposed. Furthermore, U. Scherhag *et al*. proposed new metrics including mated morph presentation match rate (MMPMR) and relative morph match rate (RMMR) to evaluate biometric systems' vulnerability to morphing attacks [5]. After that, some other metrics including attack presentation classification error rate (APCER) and the bona fide presentation classification error rate (BPCER) are complemented to evaluate the biometric systems' vulnerability to morphing attacks [6]. In addition, the vulnerability of deep learning based FRS to morphing attack was investigated in [7], and it is found that morphing attacks can degrade the performance of FRS to a certain extent.

### B. FACE MORPHING DETECTION

The existing face morphing methods can be divided into two categories depending on whether the attack accomplice's facial image can be restored or not, and they are non-restored methods and restored methods. Most face morphing detection methods belong to the first category.

#### 1) NON-RESTORED METHODS

Considering the texture discrepancy between morphed facial image and real facial image, the first work on automated face morphing detection was proposed in [8]. The micro-texture pattern represented by binarized statistical image features (BSIF) are extracted for detection. With linear support vector machine (SVM) classification, it can achieve good detection performance. As the morphed facial image is usually generated by the existing real facial images and stored in JPEG format, it will cause image quality degradation and JPEG artifact. Based on this fact, a morphed face detection method based on JPEG compression feature was proposed in [9,10], where the Benford feature extracted from quantized DCT coefficients is utilized to detect the morphed facial image. A new modeling approach for face morphing attacks was proposed by C. Kraetzer *et al.* [11]. Based on the modeling approaches, two different types face morphing attacks as well as a forensic morphing detector are implemented and evaluated. Eight features including scale invariant feature transform (SIFT), speed up robust feature (SURF), oriented BRIEF, features from accelerated segment test (FAST), adaptive and generic accelerated segment test (AGAST), Canny and Sobel edge operators are selected for the detector. At the same time, a detection approach for face morphing forgeries based on a continuous image JPEG degradation was proposed by T. Neubert [12]. It is similar to the method in [11], but only SIFT, AGAST and Shi-Tomasi features are used.

In [13], deep convolutional neural network (D-CNN) features were proposed to detect both digital and print-scanned morphed facial images. Meanwhile, a morphing attack detection approach based on deep learning was proposed in [14]. Three current widely-used network architectures are investigated, and it is found that the pre-trained VGG19 network can achieve the best performance among them.

Recently, L. Debiasi *et al.* utilized some statistical characteristics of sensor pattern noise (SPN) spectrum histogram (peak position, peak value, and the product of peak position and peak value) to detect face morphing attack [15]. It can achieve a fairy good detection performance. At the same time, illuminated by the idea of image source identification, a face morphing detection method based on SPN was put forward by us [16]. The morphed facial image is viewed as a computer-generated image, and its SPN is different from that of the real facial image.

2) RESTORED METHODS

Different from non-restored methods, restored methods usually need an extra auxiliary image captured from FRS. M. Ferrara *et al.* [17] first proposed a method for restoring the morphing attack accomplice's facial image, which is named as "face de-morphing". With an auxiliary image captured from FRS and a morphed image stored in the e-passport, it can restore the facial image of the accomplice by reversing the morphing process. However, it needs some prior knowledge about the morphing operation process and the morphing parameters of the face morphing. Once an attacker uses other morphing operations or the morphing parameters are different from the recommended morphing parameters in [17], the restored effect will be greatly affected.

From the above analysis, face morphing detection is an emerging technology to protect biometrics. Although some work has been done to detect the existence of face morphing attack, there still has no reliable approach for restoring the facial image of the morphing accomplice.

In recent years, GAN [18] has been successfully applied in various face synthesis related tasks such as photorealistic frontal synthesis [19], anti-makeup [20], and identity preserving [21]. GAN based face synthesis shows powerful advantages in realistic face generation. To countermeasure the defects of the existing methods for restoring the morphing attack accomplice's facial image, a face de-morphing generative adversarial network FD-GAN is proposed for restoring accomplice's facial image from the perspective of learning-based generation in this paper.

### III. The proposed FD-GAN

To restore the facial image of a morphing accomplice, two input facial images are required. One is an image $I_a$ of a criminal captured by an FRS, and the other is a morphed facial image $I_{ab}$ synthesized from a criminal and an accomplice, which is stored in the e-passport system. By using a symmetric dual network architecture, FD-GAN can effectually separate the morphing accomplice's identity feature $f_b$ from $I_{ab}$, and the accomplice's facial image $I_b^1$ can be restored by the generator. In addition, an adversarial discriminators $D$ is designed to distinguish between the real pair images $\{I_b^0, I_b^0\}$ and the fake (restored) pair images $\{I_b^0, I_b^1\}$. The structure of the proposed FD-GAN is depicted in Figure 2, and the symmetric dual network architecture, generator and discriminator are described in the following.

#### A. THE SYMMETRIC DUAL NETWORK ARCHITECTURE

Different from the previous GAN-based identity preserved face synthesis, the target of this paper is to restore the accomplice's facial image $I_b^1$ with a morphed facial image $I_{ab}$ and the criminal's facial image $I_a$. Here, a symmetric dual network architecture is proposed to accomplish the separation of the morphing contributor's identity.

As a morphed facial image is generally a synthesis of two participants with different identities (e.g. a criminal and an accomplice), it is reasonable to assume that the morphed image contains the identity characteristics of two contributors at the same time, so the morphing contributors' identity recovery operations should be symmetrical. It means that if the accomplice's identity feature $f_b$ can be separated from a criminal's facial image $I_a$ and a morphed image $I_{ab}$, the criminal's identity feature $f_a$ also can be separated from the same face morphing image $I_{ab}$ and an accomplice's facial image $I_b$.

Therefore, the restoration of the morphing contributors should have a symmetrical network structure. Unlike Cycle-GAN [22], which is used for domain-to-domain image style conversion by two reciprocal generators, our FD-GAN does not deal with the conversion problem from one image domain to another, and it has only one generator.

Furthermore, in order to guarantee the efficacy of the separated identity features, a dual network is intuitively considered. Assuming that once an accomplice's facial image $I_b^1$ is restored from $I_a$ and $I_{ab}$, if this restoration is valid, $I_b^1$ can be further used to restore the criminal's facial image $I_a^2$. In other words, with this dual architecture, the latter-level generation network can be used to verify the effectiveness of the previous-level generation network, and guide the previous-level generation network to separate the identity in the right way.

facial image $I_b^1$ can be obtained by facial restoration network $R$.

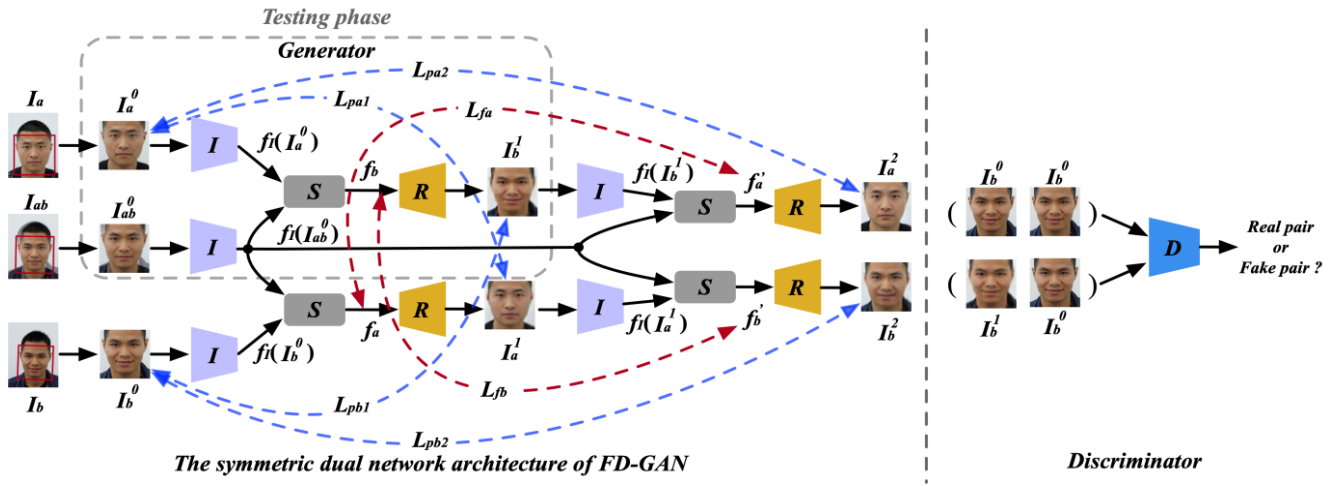

**FIGURE 2.** The overview of FD-GAN. The input/output are drawn with solid lines. The loss functions are drawn with the dashed lines. Blue dashed lines represent pixel-level restoration loss, and red dashed lines represent feature-level restoration loss. Network modules with the same color indicate that they have the same network structure and share parameters in the training stage.

Based on the above, FD-GAN with a symmetric dual network architecture is designed, and it is depicted in details in Figure 2.

### B. GENERATOR ARCHITECTURE AND GENERATOR LOSS

In FD-GAN, the generator aims to restore a desirable morphing accomplice's facial image from a morphed facial image and a criminal's facial image. The generator contains three parts: 1) an identity encoder network $I$; 2) an identity separation network $S$; and 3) a facial restoration network $R$.

Given a criminal's facial image $I_a$ and a morphed facial image $I_{ab}$, face alignment is first performed to them, and then the identity encoder network $I$ is used to extract the identity encoder features $f_I(I_a^0)$ and $f_I(I_{ab}^0)$, respectively. It needs to mention that the identity encoder feature is different from the restored identity feature, e.g. $f_I(I_b^0)$ is different from $f_b$ in Figure 2. After that, two identity encoder features are sent to the identity separation network $S$, which is used to separate the collaborator's identity feature. The structure of $S$ is shown in Figure 3. The identity coding features $f_I(I_a^0)$ and $f_I(I_{ab}^0)$ are passed through three residual blocks [23], respectively, and then they are concatenated. After another three residual blocks, the restored identity feature $f_b$ is obtained. Finally, the restored

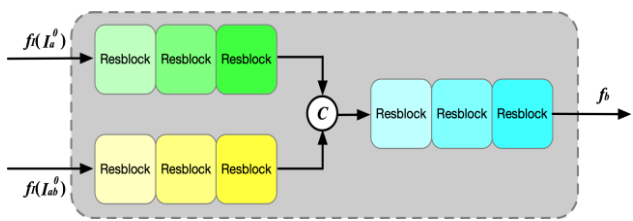

**FIGURE 3.** The structure of the identity separation network S.

TABLE I
THE ARCHITECTURE OF IDENTITY ENCODER NETWORK I

| Group Name | Configuration (filter/stride) |
|---|---|
| Conv1 | 7 × 7/1 Conv@64 + BN + ReLU |
| Conv2 | 3 × 3/2 Conv@128 + BN + ReLU |
| Conv3 | 3 × 3/2 Conv@256 + BN + ReLU |
| Conv4 | 3 × 3/2 Conv@512 + BN + ReLU |

TABLE II
THE ARCHITECTURE OF FACIAL RESTORATION NETWORK R

| Group Name | Configuration (filter/stride) |
|---|---|
| Upconv1 | Upsampling + 3 × 3/1 Conv@256 + BN + ReLU |
| Upconv2 | Upsampling + 3 × 3/1 Conv@128 + BN + ReLU |
| Upconv3 | Upsampling + 3 × 3/1 Conv@64 + BN + ReLU |
| Upconv4 | 7 × 7/1 Conv@3 + Tanh |

The details of the identity encoder network $I$ and the facial restoration network $R$ are shown in Table I and Table II, respectively. For the identity encoder network $I$, four convolution layers are used. Similar to DC-GAN [27], the pooling layer are replaced with a strided convolution layer. For the facial restoration network $R$, it is an inverse structure of $I$. The up-sampling layers are used, and the deconvolution layers are replaced with convolution layers. The "up-sampling + convolution layer" structure can effective suppress the "checkerboard" artifact of the traditional de-convolution layer in generating images. The ReLU activation is used in network $I$, $S$, and $R$.

The generator receives three kinds of losses for updating parameters: pixel-level restoration loss $L_{pxl}$, feature-level restoration loss $L_f$ and adversarial loss $L_{adv}$. The generator's total loss is

$$L_G = L_{pxl} + \lambda_1 L_f + \lambda_2 L_{adv}, \quad (1)$$

where $\lambda_1, \lambda_2$ represent the weights of the corresponding losses, respectively.

1) PIXEL-LEVEL RESTORATION LOSS

The pixel-level restoration loss $L_{pxl}$ is composed of two parts: pixel-wise loss $L_{pix}$ and symmetry loss $L_{sym}$, and it is defined as

$$L_{pxl} = L_{pix} + \beta_1 L_{sym}, \quad (2)$$

where $\beta_1$ represents the weight of symmetry loss $L_{sym}$.

Pixel-wise loss $L_{pix}$ continuously pushes the restored facial image to the ground truth as close as possible. Considering that our proposed FD-GAN uses a symmetric dual network architecture during training stage, there are four pairs of pixel-wise losses $L_{pb1}, L_{pb2}, L_{pa1},$ and $L_{pa2}$. They are defined as

$$L_{pix} = L_{pb1} + L_{pb2} + L_{pa1} + L_{pa2}, \quad (3)$$

$$L_{pb1} = \left\| I_b^1 - I_b^0 \right\|_1, \quad (4)$$

$$L_{pb2} = \left\| I_b^2 - I_b^0 \right\|_1, \quad (5)$$

$$L_{pa1} = \left\| I_a^1 - I_a^0 \right\|_1, \quad (6)$$

$$L_{pa2} = \left\| I_a^2 - I_a^0 \right\|_1, \quad (7)$$

where $I_b^1 = G(I_a^0, I_{ab}^0)$, $I_a^1 = G(I_b^0, I_{ab}^0)$, $I_b^2 = G(G(I_b^0, I_{ab}^0), I_{ab}^0)$, $I_a^2 = G(G(I_a^0, I_{ab}^0), I_{ab}^0)$, $G(\cdot, \cdot)$ represents the generator of FD-GAN, and $\|\bullet\|_1$ represents $L_1$-norm.

As symmetry is a prominent characteristic of human face, it is taken into account as a constraint to guarantee the rationality of the restored face. Similarly, there also have four symmetry losses $L_{symI_b^1}, L_{symI_b^2}, L_{symI_a^1},$ and $L_{symI_a^2}$, and they are defined as

$$L_{sym} = L_{symI_b^1} + L_{symI_b^2} + L_{symI_a^1} + L_{symI_a^2}, \quad (8)$$

$$L_{symI_b^1} = \frac{1}{h \times w/2} \sum_{i=1}^{h} \sum_{j=1}^{w} \left\| (I_b^1)_{i,j} - (I_b^1)_{i,w-j+1} \right\|_1, \quad (9)$$

$$L_{symI_b^2} = \frac{1}{h \times w/2} \sum_{i=1}^{h} \sum_{j=1}^{w} \left\| (I_b^2)_{i,j} - (I_b^2)_{i,w-j+1} \right\|_1, \quad (10)$$

$$L_{symI_a^1} = \frac{1}{h \times w/2} \sum_{i=1}^{h} \sum_{j=1}^{w} \left\| (I_a^1)_{i,j} - (I_a^1)_{i,w-j+1} \right\|_1, \quad (11)$$

$$L_{symI_a^2} = \frac{1}{h \times w/2} \sum_{i=1}^{h} \sum_{j=1}^{w} \left\| (I_a^2)_{i,j} - (I_a^2)_{i,w-j+1} \right\|_1, \quad (12)$$

2) FEATURE-LEVEL RESTORATION LOSS

Besides pixel-level restoration losses, there still has feature-level restoration loss. For the proposed symmetrical dual network architecture, the restored identity features in each stage should be as consistent as possible. There are two pairs of feature-level restoration losses $L_{fa}$ and $L_{fb}$, and they are defined as

$$L_f = L_{fa} + L_{fb}, \quad (13)$$

$$L_{fa} = \left\| f_a' - f_a \right\|_1, \quad (14)$$

$$L_{fb} = \left\| f_b' - f_b \right\|_1. \quad (15)$$

3) ADVERSARIAL LOSS

The adversarial loss of the generator for distinguishing the real pair facial images $\{I_b^0, I_b^0\}$ from the fake (restored) pair facial images $\{I_b^0, I_b^1\}$ is calculated as

$$L_{adv} = \mathbb{E}_{I_b^0 \sim P(I_b^0), I_a^0 \sim P(I_a^0), I_{ab}^0 \sim P(I_{ab}^0)}[(D(I_b^0, I_b^1) - 1)^2], \quad (16)$$

where $D(\cdot, \cdot)$ represents the discriminator of FD-GAN.

C. DISCRIMINATOR ARCHITECTURE AND LOSS

Different from the traditional GAN discriminators, the discriminator of FD-GAN determines both the real pair facial images $\{I_b^0, I_b^0\}$ and the restored pair facial images $\{I_b^0, I_b^1\}$. The details of discriminator $D$ are shown in Table III. In the network $D$, three convolution layers are used, and the Leaky ReLU activation is used in network $D$.

The discriminator requires that the restored facial image $I_b^1$ should like a real facial image, and it should belong to the same identity as $I_b^0$. Here, LS-GAN [24] loss instead of the negative log likelihood, which was used in the original GAN, is used. The discriminator loss is defined as

$$L_D = \mathbb{E}_{I_b^0 \sim P(I_b^0)}[(D(I_b^0, I_b^0) - 1)^2] + \mathbb{E}_{I_b^0 \sim P(I_b^0), I_a^0 \sim P(I_a^0), I_{ab}^0 \sim P(I_{ab}^0)}[(D(I_b^0, I_b^1))^2]. \quad (17)$$

TABLE III
THE ARCHITECTURE OF DISCRIMINATOR NETWORK D

| Group Name | Configuration (filter/stride) |
|---|---|
| Conv1 | 3 × 3/2 Conv@64 + Leaky ReLU |
| Conv2 | 3 × 3/2 Conv@128 + BN + Leaky ReLU |
| Conv3 | 3 × 3/2 Conv@256 + BN + Leaky ReLU |
| FC1 | Fully Connected Layer+ Sigmoid |

IV. Experiments and Analysis

As there is no publicly available facial image dataset for face morphing detection, we built a face morphing database (FM-database) and then it is used to evaluate the effectiveness of the proposed FD-GAN.

A. THE SETUP OF THE FM-DATABASE

The FM-database contains 63 subjects (27 females and 36 males). It is divided into three disjoint subsets: training set, development set and testing set, and they consist of 28, 7 and 28 subjects, respectively. All subjects were photographed twice with an interval of one week and 10~12 frontal images were taken each time. Thus, there are about 23 images for each of 63 subjects. One image with ideal quality of each subject is chosen for creating morphed facial images and the rest are used to construct the bona fide facial images (with a total of 1378 images). The facial images of all subjects are captured by Canon 550D SLR camera following the ICAO guidelines on eMRTD. To guarantee the disjoint nature of the database as mentioned above, the morphed facial images are independently generated in training, development, and testing set. In each subset, face morphing images are automatically generated according to the workflow proposed in [9].

To effectively evaluate the validity of the FM database, a commercial FRS Megvii Face++ Compare API [25] is exploited. A successful morphing attack to the FRS is defined that both subjects of the morphed image can be successfully accepted by the system [2]. It means that each morphed facial image needs to be compared with the images of all subjects, which are contributed to the morphing process. By using Face++, the Compare API requires to be called twice, and only two returned scores are both greater than the recommended threshold can the morphed facial image be considered to be a successful face morphing attack. The recommended threshold of Face++ Compare API is 62.37 when FAR is 0.1% (FRONTEX recommends that the operation point of FRS in border control scenarios is at FAR=0.1%).

Finally, 1378 morphed facial images (with various fusion factors) that successfully attack the Face++ are collected into the FM-database. Among them, 584 morphed facial images are used for restoration testing.

### B. IMPLEMENTATION DETAILS
In order to make the training phase effective, all the input images are aligned according to 5 facial landmarks by using dlib programming library [26]. Each facial image is resized to $128 \times 128 \times 3$ pixels, and the restored facial images from the generator are of the same size.

The FD-GAN is implemented using the deep learning toolbox Tensorflow 1.11.0 [28]. For the loss weights, they are empirically set as $\lambda_1 = 10$, $\lambda_2 = 1$ and $\beta_1 = 1$.

### C. EXPERIMENTAL RESULTS
In the experiments, two scenarios are considered. The first is the ideal scenario. The template facial images are used to generate the morphing facial images, and each of them is obtained as an auxiliary image for training and testing. The purpose is to verify the feasibility of the proposed FD-GAN. The second is the real application scenario. The facial images captured by FRS are used for training and testing, which may contain various expression, make-up and occlusion.

#### 1) ANALYSIS OF THE PROPOSED FRAMEWORK
For the first scenario, an ablation study is performed. Different loss combinations and without use of symmetric dual network are done to the FD-GAN to evaluate its performance. Here, four variations of FD-GAN are compared, and they are 1) removing the loss $L_{pix}$ (denoted as w/o $L_{pix}$); 2) removing the loss $L_f$ (denoted as w/o $L_f$); 3) removing the loss $L_{sym}$ (denoted as w/o $L_{sym}$); and 4) training without symmetric dual network architecture (denoted as w/o SDN). The visualization samples of each variant are shown in Figure 4. Where GT represents the ground truth of the accomplice $I_b^0$, and $I_b^0$, $I_{ab}^0$ represent the criminal facial image and morphed facial image, respectively. As expected, FD-GAN including all losses achieves the best visual effects. Without the loss $L_{pix}$, it is almost impossible to generate the real facial images. Without symmetric dual network architecture, the restored facial images are obviously different from the ground truth. After removing the loss $L_f$, the restored facial images are also different from the ground truth, but there is a little improvement comparing with those without symmetrical structure network. When the loss $L_{sym}$ is removed, the facial asymmetry is apparent in nostrils and eyes' size.

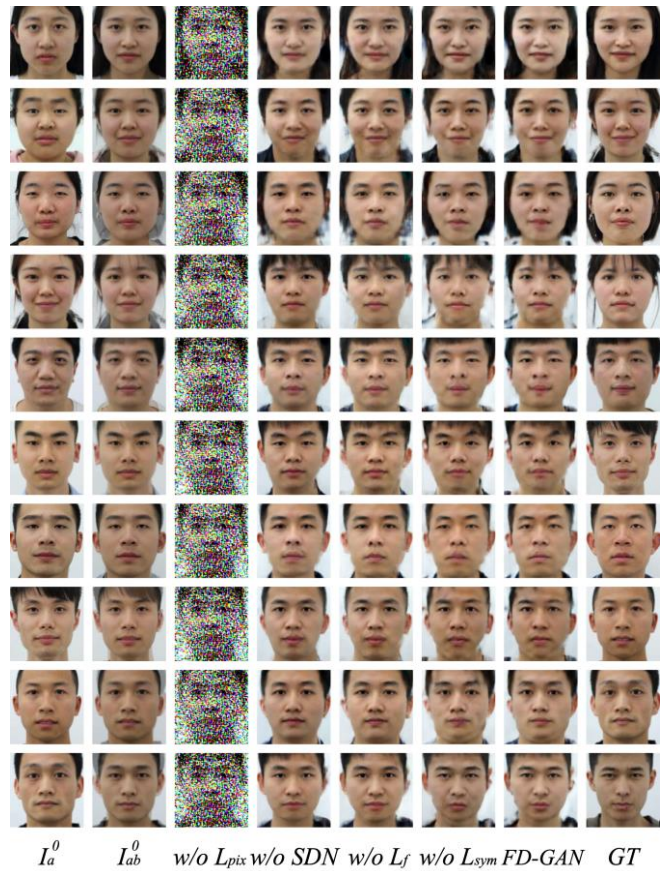

**FIGURE 4.** Restoration results of FD-GAN and its variants.

To quantitatively evaluate the performance of FD-GAN, Face++ is used to compare the restored facial image $I_b^1$ with $I_b^0$ and $I_a^0$, respectively. When the system determines that $I_b^1$ matches $I_b^0$ but it does not match $I_a^0$, the restoration is regarded as successful. The restoration accuracy is defined as

$$Accuracy = \frac{N}{T}, \quad (18)$$

where $N$ represents the number of successfully restored facial images, and $T$ represents the total number of restored facial images.

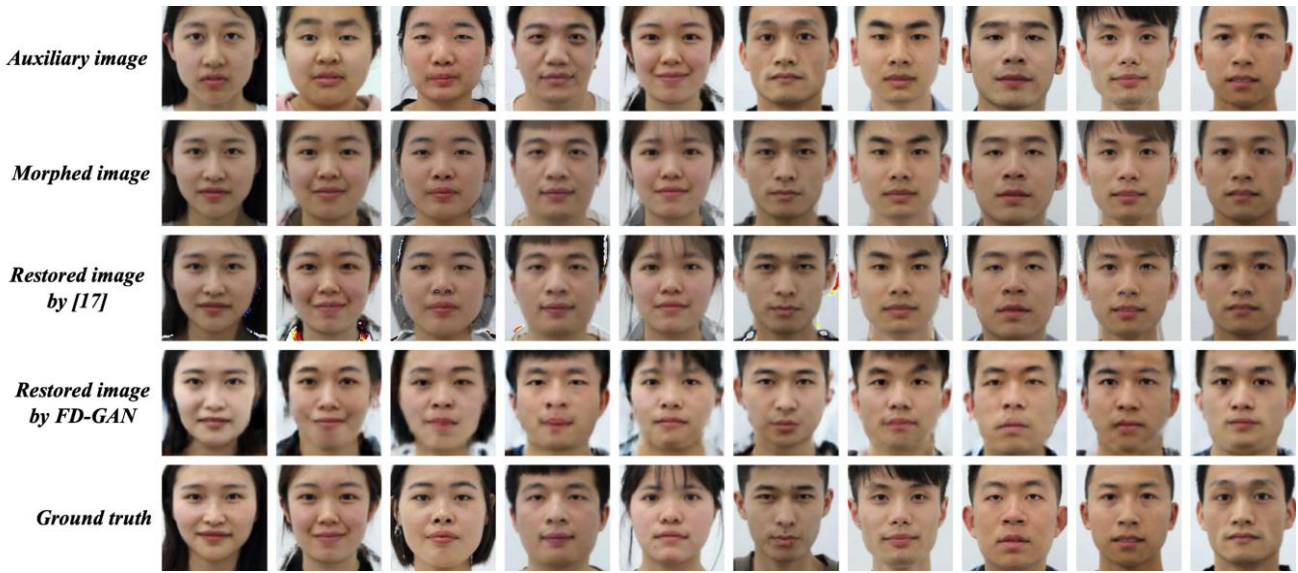

**FIGURE 5.** Example of restoration comparison in scenario 1.

TABLE IV
THE RESTORATION ACCURACY WITH ABLATION IN SCENARIO 1

| Method | w/o $L_{pix}$ | w/o SDN | w/o $L_f$ | w/o $L_{sym}$ | FD-GAN |
|---|---|---|---|---|---|
| Accuracy | - | 45.55% | 49.32% | 79.11% | **85.97%** |

TABLE V
THE RESTORATION ACCURACY WITH ABLATION IN SCENARIO 2

| Method | w/o $L_{pix}$ | w/o SDN | w/o $L_f$ | w/o $L_{sym}$ | FD-GAN |
|---|---|---|---|---|---|
| Accuracy | - | 33.56% | 49.82% | 61.13% | **64.90%** |

TABLE VI
THE RESTORATION ACCURACY COMPARISON IN TWO SCENARIOS

| Method | Scenario 1 | Scenario 2 |
|---|---|---|
| Face de-morphing [17] | 49.82% | 46.91% |
| FD-GAN (Proposed) | **85.97%** | **64.90%** |

The restoration accuracy of scenario 1 is listed in Table IV. For the second scenario, the same hyper-parameters (loss weight and learning rate etc.) of the first scenario are used for simplicity. Some synthetic samples are illustrated in Figure 6, and the restoration accuracy is listed in Table V.

Comparing the results listed in Table IV and Table V, the restoration accuracy of the second scenario is much lower than that of the first situation. This is mainly because the facial images captured by FRS in the second situation have more variants in expression, make-up, occlusion, etc. Therefore, it is more difficult to restore the face morphing accomplice.

### 2) PERFORMANCE COMPARISON

In order to evaluate the effectiveness of the proposed scheme, the proposed FD-GAN is also compared with the latest face de-morphing method [17] in two scenarios (The de-morphing parameter was selected as 0.3, which is recommended in [17]). The restoration accuracy is listed in Table VI, and some synthetic samples are also illustrated in Figure 5 and Figure 6.

From Table VI, it can be found that the accuracy of restoring accomplice's facial image of the proposed method outperforms that of the method in [17]. In scenario 1, FD-GAN achieve the best restored accuracy 85.97%, while the method [17] is only 49.82%. As seen from Figure 5 and Figure 6, the restored facial images with the method in [17] have different degrees of restoration defects, such as abnormal color spots, edge artifacts, etc. The most obvious is that in the last four examples in Figure 5. The restored facial images are almost the same as the morphed facial images, and the facial images of the accomplices are not effectively recovered. The main reason is that the method in [17] relies on the prior knowledge of the generation of the morphed face, such as the morphing process and the morphing parameters. In this experiment, the varied morphing parameters are used in morphed facial image generation, which leads to limited restoration performance. It also indicates that the method in [17] is sensitive to the prior knowledge and not suitable for practical applications. However, the proposed FD-GAN doesn't rely on any prior knowledge of the generation of the morphed face, and it realizes the accomplice's facial restoration by separating the identity of the accomplice hidden in morphed facial images.

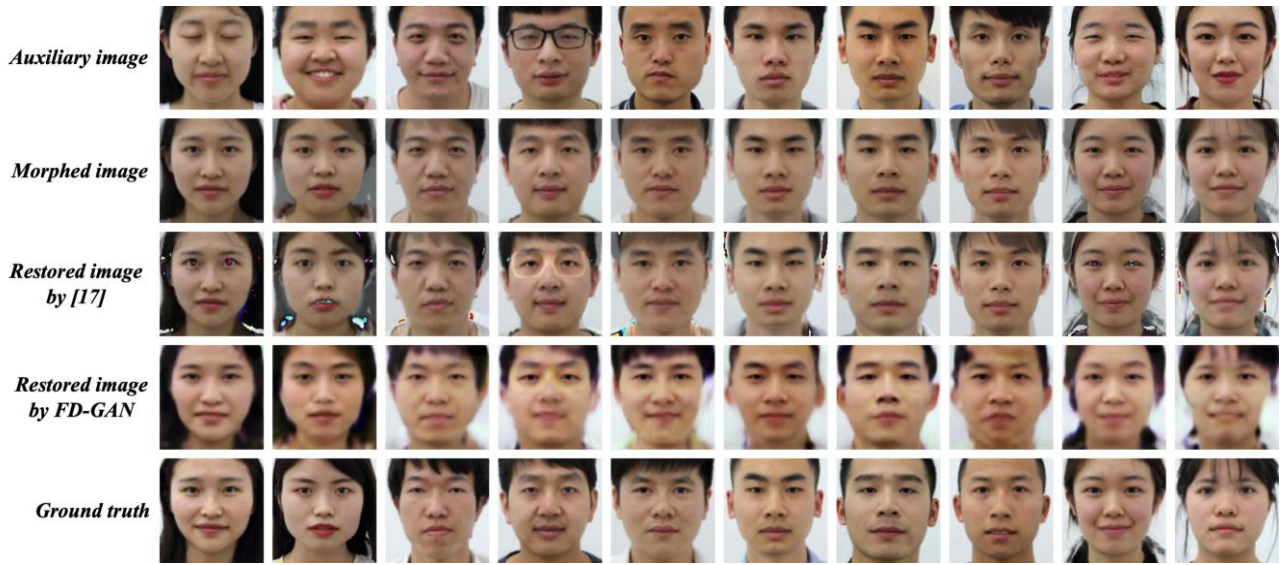

**FIGURE 6.** Example of restoration comparison in scenario 2 (auxiliary images contain expression, make-up and occlusion changes).

The experimental results show the superiority of the proposed FD-GAN of its symmetric dual structure. In addition, since all morphed images used in the experiments can successfully spoof Face++, the implementation of FD-GAN can significantly reduce the possibility of successful face morphing attack and further restore the accomplice's facial image.

## V. CONCLUSION

In this paper, a FD-GAN is proposed to restore the face morphing accomplice's facial image. To the best of our knowledge, it is the first work to restore the face morphing accomplice's facial image using learning-based generation approach. The symmetric dual network is specially designed to effectively disentangle the identity features of the morphing participants. Experimental results demonstrate that the proposed FD-GAN can achieve fairly good restoration performance. Our future work will be concentrated on improving the morphing contributor's facial restoration accuracy involving expression, posture, make-up and occlusion variants.

## APPENDIX

### AUTOMATIC FACE MORPHING PROCESS

According to the workflow of [1], the general steps for automatic generation of the morphed facial image are as follows.

*Step* **1**. Locate the key points of the morphed facial image. Given two facial images $I_1$ and $I_2$, $K_1$ and $K_2$ are the locations of the corresponding facial landmarks. Then the location $K_M$ of the key points of the morphed facial image can be calculated as

$$K_M = (1-\beta)K_1 + \beta K_2, \quad (19)$$

where $\beta$ represents a location fusion factor.

To avoid the failure of morphing face generation, 85 landmarks are used for generating morphed facial image, as shown in Figure 7. Among them, 68 facial landmarks are localized by using dlib programming library [26]. However, three facial landmarks depicting the lower contour of the upper lip are omitted, because they may result in morphing errors. The other 20 landmarks added on the image borders are used for triangulation in the outer region of the convex hull, which is composed by 65 facial landmarks. With the help of facial landmark points of both contributing images, a morphed facial image using triangulation based tight morphing is generated.

*Step* **2**. Warping. Triangular mesh $T_M$ can be derived from $K_M$ via Delaunay triangulation. Similarly, two triangular mesh $T_1$ and $T_2$ can also be derived from $K_1$ and $K_2$, respectively. For each triangle in $T_1$, all pixels inside the triangle are transformed to the corresponding triangle in $T_M$ by affine transform, and then the warped image $I_1'$ from $I_1$ is obtained. Similarly, the warped image $I_2'$ from $I_2$ can be also obtained.

Step 3. Blending. A morphed facial image $I_M$ can be obtained as

$$I_M = (1-\alpha)I_1' + \alpha I_2', \quad (20)$$

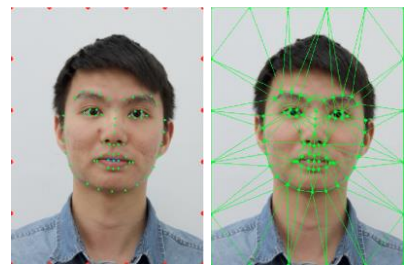

**FIGURE 7.** Example of facial landmarks and corresponding Delaunay triangles (green and blue landmarks are localized by dlib, blue landmarks are removed, and red points are additional landmarks (red).

where $\alpha$ represents a pixel fusion factor.

In the experiments, the pixel fusion factor $\alpha$ is varied from 0.1 to 0.9, and the location fusion factor $\beta$ is fixed as 0.5.

## ACKNOWLEDGMENT
This work was supported in part by project supported by National Natural Science Foundation of China (Grant Nos. 61572182, 61370225).

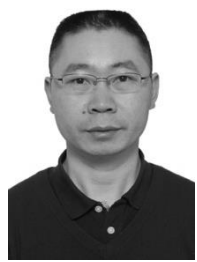

**Fei Peng** received the Ph.D. degree in Circuits and Systems from the South China University of Science and Technology, Guangzhou, China, in 2006. He was a visiting fellow of the Department of Computer Science, University of Warwick, U.K. in 2009–2010. He was a visiting professor in SeSaMe Centre, College of Computing, National University of Singapore in 2016. Currently, he is a professor in the College of Computer Science and Electronic Engineering, Hunan University, Changsha. His areas of interest include digital watermarking and digital forensics.

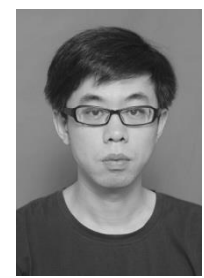

**Le-Bing Zhang** received the B.S. and M.S. degrees from Hunan Normal University in 2003 and 2009, respectively. Currently, He is a Ph.D. candidate of the College of Computer Science and Electronic Engineering, Hunan University, Changsha. He is also an assistant professor with Huaihua University, Huaihua, China. His areas of interest include digital forensics, multimedia security and deep learning.

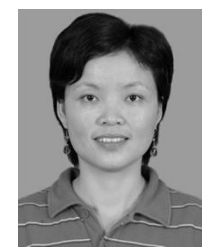

**Min Long** received the Ph.D. degree in Circuits and Systems from the South China University of Science and Technology, Guangzhou, China, in 2006. She was a visiting fellow of the Department of Computer Science, University of Warwick, U.K. in 2009–2010. Currently, she is a professor in the College of Computer and Communication. Her areas of interest include digital watermarking and chaos-based secure Communication.